\documentclass{article}
\usepackage{ijcai21}
\usepackage{times}
\usepackage{soul}
\usepackage{url}
\usepackage[hidelinks]{hyperref}
\usepackage[utf8]{inputenc}
\usepackage[small]{caption}
\usepackage{graphicx}
\usepackage{amsmath}
\usepackage{amsthm}
\usepackage{booktabs}
\usepackage{algorithm}
\usepackage{algorithmic}
\usepackage{amssymb}

\title{Deep Automatic Natural Image Matting}

\author{
Jizhizi Li$^1$\and
Jing Zhang$^1$\And
Dacheng Tao$^{2}$\\
\affiliations
$^1$The University of Sydney, Australia\\
$^2$JD Explore Academy, JD.com, China\\
\emails
jili8515@uni.sydney.edu.au,
jing.zhang1@sydney.edu.au,
dacheng.tao@gmail.com
}

\begin{document}

\maketitle


\begin{abstract}
Automatic image matting (AIM) refers to estimating the soft foreground from an arbitrary natural image without any auxiliary input like trimap, which is useful for image editing. Prior methods try to learn semantic features to aid the matting process while being limited to images with \textit{salient opaque} foregrounds such as humans and animals. In this paper, we investigate the difficulties when extending them to natural images with \textit{salient transparent/meticulous} foregrounds or \textit{non-salient} foregrounds. To address the problem, a novel end-to-end matting network is proposed, which can predict a generalized trimap for any image of the above types as a unified semantic representation. Simultaneously, the learned semantic features guide the matting network to focus on the transition areas via an attention mechanism. We also construct a test set AIM-500 that contains 500 diverse natural images covering all types along with manually labeled alpha mattes, making it feasible to benchmark the generalization ability of AIM models. Results of the experiments demonstrate that our network trained on available composite matting datasets outperforms existing methods both objectively and subjectively. The source code and dataset are available at \href{https://github.com/JizhiziLi/AIM}{https://github.com/JizhiziLi/AIM}.
\end{abstract}


\section{Introduction}
\label{sec:introduction}

\begin{figure}[t]
    \includegraphics[width = 1\linewidth]{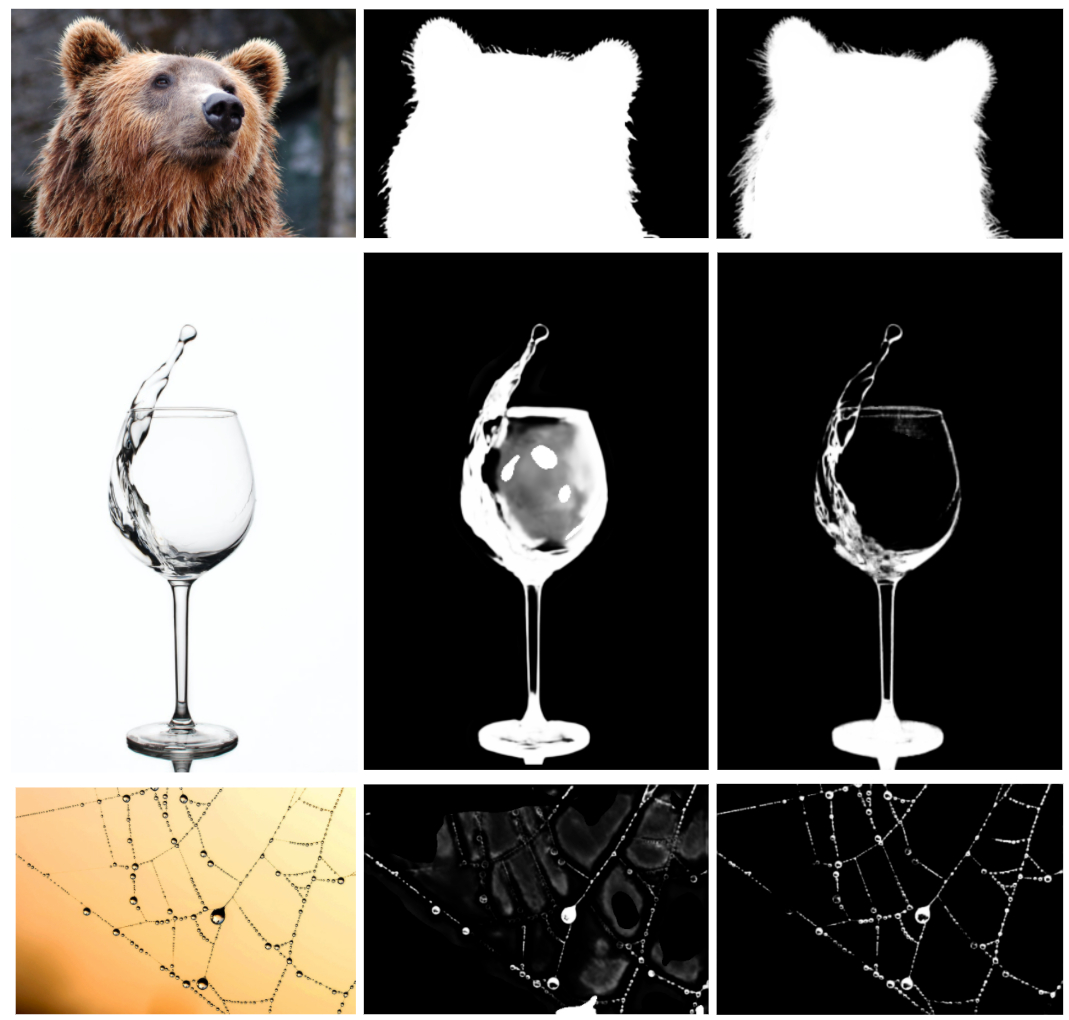}
    \caption{Matting results of GFM and our method on three types of natural images, i.e., SO, STM, NS, from the top to the bottom.}
    \label{fig:introduction}
\end{figure}

Natural image matting refers to estimating a soft foreground from a natural image, which is a fundamental process for many applications, e.g., film post-production and image editing~\cite{chen2013knn,zhang2020empowering}. Since image matting is a highly ill-posed problem, previous methods usually adopt auxiliary user input, e.g. trimap~\cite{sun2004poisson,cai2019disentangled}, scribble~\cite{levin2007closed}, or background image~\cite{backgroundmatting} as constraints. While traditional methods estimate the alpha value by sampling neighboring pixels~\cite{wang2007optimized} or defining affinity metrics for alpha propagation~\cite{levin2008spectral}, deep learning-based approaches solve it by learning discriminative representations from a large amount of labeled data and predicting alpha matte directly~\cite{lu2019indices,li2020natural}.

However, extra manual effort is required while generating such auxiliary inputs, which makes these methods impractical in automatic industrial applications. To address the limitations, automatic image matting (AIM) has attracted increasing attention recently~\cite{zhang2019late}, which refers to automatically extracting the soft foreground from an arbitrary natural image. Prior AIM methods~\cite{Qiao_2020_CVPR,gfm} solve this problem by learning semantic features from the image to aid the matting process but is limited to images with salient opaque foregrounds, e.g., human~\cite{shen2016deep,chen2018semantic}, animal~\cite{gfm}. It is difficult to extend them to images with salient transparent/meticulous foregrounds or non-salient foregrounds due to their limited semantic representation ability. 

To address this issue, we dig into the problem of AIM by investigating the difficulties of extending it to all types of natural images. First, we propose to divide natural images into three types according to the characteristics of foreground alpha matte. As shown in Figure~\ref{fig:introduction}, the three types are: 1) \textbf{SO} (Salient Opaque): images that have salient foregrounds with opaque interiors, e.g. human, animal; 2) \textbf{STM} (Salient Transparent/Meticulous): images that have salient foregrounds with transparent or meticulous interiors, e.g. glass, plastic bag; and 3) \textbf{NS} (Non-Salient): images with non-salient foregrounds, e.g. smoke, grid, raindrop. Some examples are shown in Figure~\ref{fig:introduction}. Then, we systematically analyze the ability of baseline matting models for each type of image in terms of understanding global semantics and local matting details. We find that existing methods usually learn implicit semantic features or use an explicit semantic representation that is defined for a specific type of image, i.e., SO. Consequently, they are inefficient to handle different types of images with various characteristics in foreground alpha mattes, e.g., salient opaque/transparent and non-salient foregrounds.

In this paper, we make the first attempt to address the problem by devising a novel automatic end-to-end matting network for all types of natural images. First, we define a simple but effective unified semantic representation for the above three types by generalizing the traditional trimap according to the characteristics of foreground alpha matte. Then, we build our model upon the recently proposed effective GFM model \cite{gfm} with customized designs. Specifically, 1) we use the generalized trimap as the semantic representation in the semantic decoder to adapt it for all types of images; 2) we exploit the effective SE attention \cite{hu2018squeeze} in the semantic decoder to learn better semantic features to handle different characteristics of foreground alpha mattes; 3) we improve the interaction between the semantic decoder and matting decoder by devising a spatial attention module, which guides the matting decoder to focus on the details in the transition areas. These customized designs prove to be effective for AIM rather than trivial tinkering, as shown in Figure~\ref{fig:introduction}.

Besides, there is not a test bed for evaluating AIM models on different types of natural images. Previous methods use composition images by pasting different foregrounds on background images from  COCO dataset~\cite{lin2014microsoft} for evaluation, which may introduce composition artifacts and are not representative for natural images as mentioned in~\cite{gfm}. Some recent works collect natural test images with manually labeled alpha mattes, however, they are limited to specific types of images, such as portrait images~\cite{MODNet} or animal images~\cite{gfm}, which are not suitable for comprehensive evaluation of matting models for AIM. To fill this gap, we establish a benchmark AIM-500 by collecting 500 diverse natural images covering all three types and many categories and manually label their alpha mattes.
 
The main contributions of this paper are threefold: 1) we make the first attempt to investigate the difficulties of automatic natural image matting for all types of natural images; 2) we propose a new matting network with customized designs upon a reference architecture that are effective for AIM on different types of images; and 3) we establish the first natural images matting benchmark AIM-500 by collecting 500 natural images covering all three types and manually labeling their alpha mattes, which can serve as a test bed to facilitate future research on AIM.


\section{Rethinking the Difficulties of AIM}

\label{sec:rethinking}

\paragraph{Matting without auxiliary inputs.}Prevalent image matting methods~\cite{sim,cho2016natural} solve the problem by leveraging auxiliary user inputs, e.g. trimap~\cite{tang2019learning}, scribble~\cite{levin2007closed}, or background image~\cite{backgroundmatting}, which provide strong constraints on the solution space. Specifically, given a trimap, the matting models only need to focus on the transition area and distinguish the details by leveraging the available foreground and background alpha matte information. However, there is usually little chance to obtain auxiliary information in real-world automatic application scenarios. Thereby, AIM is more challenging since the matting models need to understand the holistic semantic partition of foreground and background of an image, which may belong to different types as been described previously. Nevertheless, AIM is more appealing to automatic applications and worth more research efforts.

\paragraph{Matting on natural images.}Since it is difficult to label accurate alpha mattes for natural images at scale, there are no publicly available large-scale natural image matting datasets. Usually, foreground images are obtained by leveraging chroma keying from images captured with a green background screen \cite{xu2017deep}. Nevertheless, the amount of available foregrounds is only about 1,000. To build more training images, they are synthesized with different background images from public datasets like COCO \cite{lin2014microsoft}. However, synthetic images contain composite artifacts and semantic ambiguity~\cite{gfm}. Matting models trained on them may have a tendency to find cheap features from these composite artifacts and thus overfit on the synthetic training images, resulting in a poor generalization performance on real-world natural images. To address this domain gap issue between synthetic training images and natural test images, some efforts have been made in \cite{gfm,hou2019context}. As for evaluation, previous methods~\cite{zhang2019late,Qiao_2020_CVPR} also adopt synthetic dataset~\cite{xu2017deep} for evaluating AIM models, which is a bias evaluation setting. For example, a composite image may contain multiple objects including the original foreground object in the candidate background image, but the ground truth is only the single foreground object from the foreground image. Besides, the synthetic test images may also contain composite artifacts, making it less possible to reveal the overfitting issue. 

\paragraph{Matting on all types of images.}AIM aims to extract the soft foreground from an arbitrary natural image, which may have foreground objects with either opaque interior, transparent/meticulous interior, or non-salient foregrounds like texture, fog, or water drops. However, existing AIM methods are limited to a specific type of images with opaque foregrounds, e.g., human~\cite{shen2016deep,chen2018semantic,zhang2019late,li2021privacypreserving,MODNet} and animals~\cite{gfm}. DAPM~\cite{shen2016deep} first generates a coarse foreground shape mask and uses it as an auxiliary input for following matting process. LateFusion~\cite{zhang2019late} predicts foreground and background separately and uses them in a subsequent fusion process. HAttMatting~\cite{Qiao_2020_CVPR} predicts foreground profile and guides the matting process to refine the precise boundary. GFM~\cite{gfm} predicts foreground, background, and transition area simultaneously and combines them with the matting result at the transition area as the final matte. Although they are effective for images with salient and opaque foregrounds, extending them to all types of images is not straightforward. In the context of AIM, the term ``semantic'' is more related to foreground and background rather than the semantic category of foreground objects. Since there are very different characteristics for the three types of images, it is hard to learn useful semantic features to recognize foreground and background, especially for images with transparent objects or non-salient foregrounds, without explicit supervisory signals.


\section{Proposed Methods}

\subsection{Improved Backbone for Matting}

In this work, we choose ResNet-34~\cite{he2016deep} as the backbone network due to its lightweight and strong representation ability. However, ResNet-34 is originally designed to solve high-level classification problem, while both high-level semantic features and low-level detail features should be learned to solve AIM. To this end, we improve the vanilla ResNet-34 with a simple customized modification to make it better suitable for the AIM task. 

In the first convolutional layer conv1 in ResNet-34, they use a stride of 2 to reduce the spatial dimension of feature maps. Followed by a max-pooling layer, the output dimension is $1/4$ of the original image after the first block. While these two layers can significantly reduce computations and increase the receptive field, it may also lose many details, which are not ideal for image matting. To customize the backbone for AIM, we modify the stride of conv1 from 2 to 1 to keep the spatial dimension of the feature map as the original image size to retain local detail features. To retain the receptive field, we add two max-pooling layers with a stride of 2. Besides, we change the stride of all the first convolutional layers in stage1-stage4 of ResNet-34 from 2 to 1 and add a max-pooling layer with a stride of 2 accordingly. It is noteworthy that the indices in the max-pooling layers are also kept and will be used in the corresponding max-unpooling layers in the local matting decoder to preserve local details, as shown in Figure~\ref{fig:network}. We retrain the customized ResNet-34 on ImageNet and use it as our backbone network. Experiment results demonstrate it outperforms the vanilla one in terms of both objective metrics and subjective visual quality. 

\subsection{Unified Semantic Representation}

\begin{figure}[hbt!]
    \includegraphics[width = 1\linewidth]{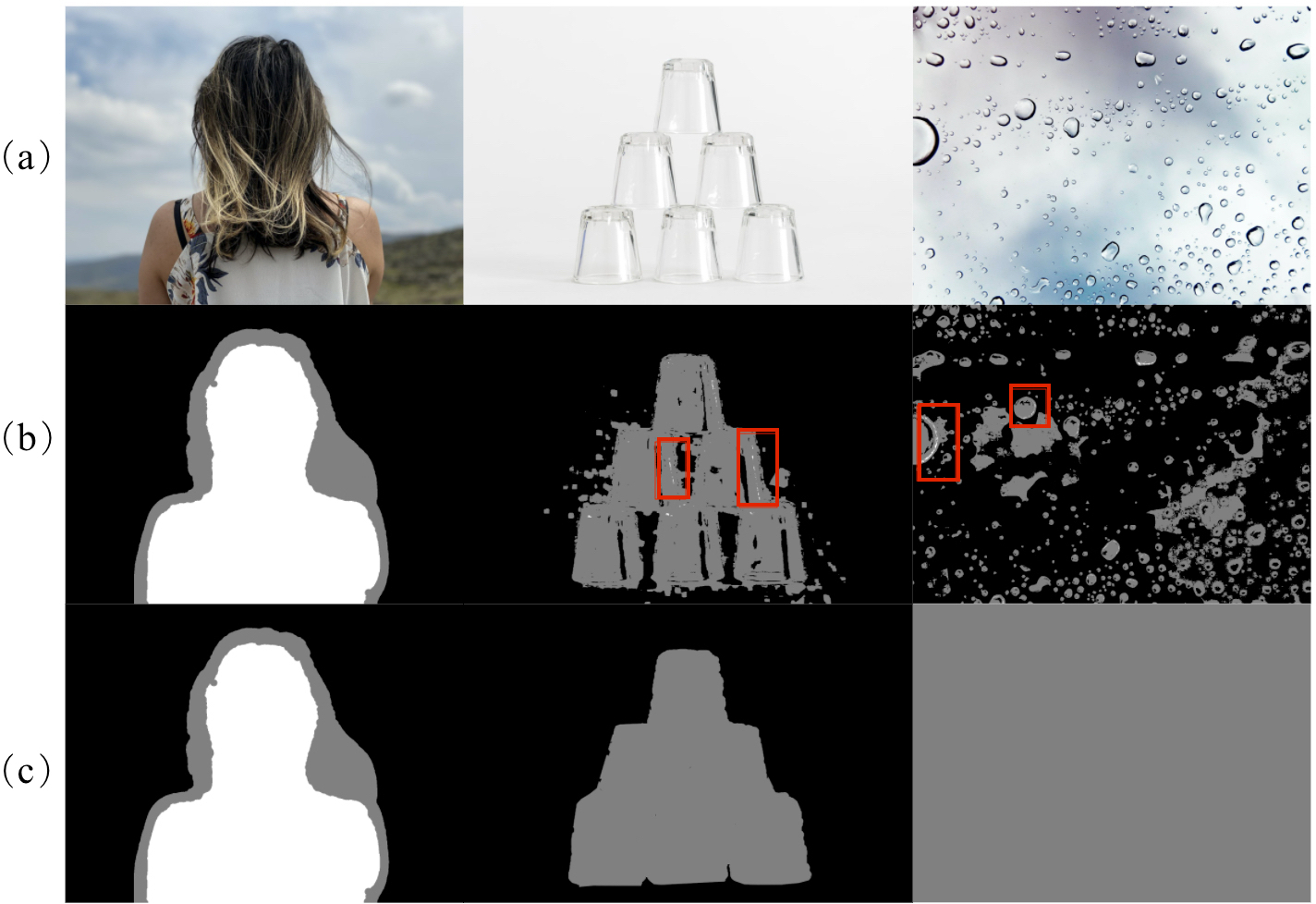}
    \caption{(a) Three types of images. (b) The traditional trimap representation. (c) The unified semantic representations, i.e., trimap, duomap, and unimap, respectively. White: foreground, Black: background, Gray: transition area.}
    \label{fig:featuremap}
\end{figure}

\begin{figure*}[t]

    \includegraphics[width = 1\linewidth]{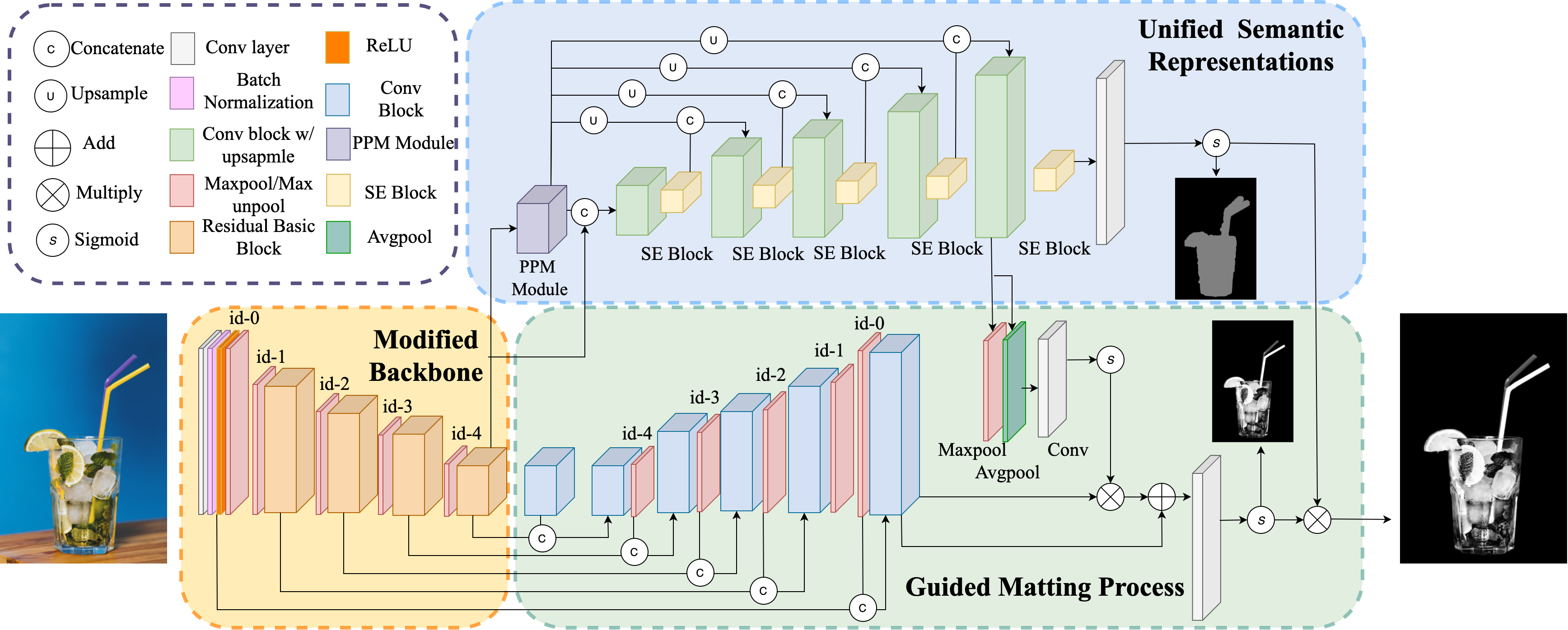}
    \caption{The structure of our matting network for AIM.}
    \label{fig:network}
\end{figure*}
As discussed in Section~\ref{sec:rethinking}, the characteristics of different types of images are very different. In order to provide explicit semantic supervisory signals for the semantic decoder to learn useful semantic features and partition the image into foreground, background, and transition areas, we propose a unified semantic representation. 

For an image belonging to SO type, there are always foreground, background, and transition areas. Thereby, we adopt the traditional trimap as the semantic representation, which can be generated by erosion and dilation from the ground truth alpha matte. For an image belonging to STM type, there are no explicit foreground areas. In other words, the foreground should be marked as a transition area and a soft foreground alpha matte should be estimated. Thereby, we use a duomap as its semantic representation, which is a 2-class map denoting the background and transition areas accordingly. For an image belonging to NS type, it is hard to mark all the explicit foreground and background areas, since the foreground is always entangled with the background. Thereby, we use a unimap as its semantic representation, which is a 1-class map denoting the whole image as the transition area. 

To use a unified semantic representation for all the three types of images, we derive the trimap, duomap, and unimap from the traditional trimap as follows:
\begin{equation}
\left\{
             \begin{array}{lr}
            U_i = T_i, &type=SO  \\
              U_i = 1.5T_i-T_i^{2} , & type=STM  \\
            U_i = 0.5, &type=NS  
             \end{array},
\right.
\label{eqa:newtri}
\end{equation}
where $T$ represents the traditional trimap obtained by erosion and dilation from the ground truth alpha matte. For every pixel $i$, $T_{i}\in \left \{ 0, 0.5, 1 \right \}$, where the background area is 0, the foreground area is 1, and the transition area is 0.5, respectively. $U_{i}$ is the unified semantic representation at pixel $i$. Note that for images belonging to type STM and NS, the traditional trimaps always contain trivial foreground and background pixels as shown in Figure~\ref{fig:featuremap}(b), which are very difficult for the semantic decoder to predict. Instead, our unified representations are a pure duomap or unimap as shown in Figure~\ref{fig:featuremap}(c), which represents the holistic foreground objects or denotes there are no salient foreground objects.

In order to predict the unified semantic representation, we redesign the semantic decoder in GFM~\cite{gfm} as shown in Figure~\ref{fig:network}. Specifically, we also use five blocks in the decoder. For each decoder block, there are three sequential $3\times3$ convolutional layers and an upsampling layer. To further increase the capability of the decoder for learning discriminative semantic features, we adopt the Squeeze-and-Excitation (SE) attention module~\cite{hu2018squeeze} after each decoder block to re-calibrate the features, thereby selecting the most informative features for predicting unified representations and filtering out the less useful ones. We also adopt pyramid pooling module (PPM)~\cite{zhao2017pyramid} to enlarge the receptive field. The upsampled PPM features are concatenated with the output of SE module and used as the input for the next decoder block. We use the cross-entropy loss to supervise the training of the semantic decoder.

\subsection{Guided Matting Process}

 As shown in Figure~\ref{fig:network}, there are six blocks following a U-Net structure~\cite{ronneberger2015u} in matting decoder.. Each block contains three sequential $3\times3$ convolutional layers. The encoder feature is concatenated with first decoder block output and fed into the following block. The output then being concatenated with the corresponding encoder output and passes a max unpooling layer with reversed indices to recover fine structural details, serves as input of next block.

Motivated by previous study proving that attention mechanism can provide support on learning discriminating representations~\cite{ma2020auto}, we devise a spatial attention module to guide the matting process by leveraging the learned semantic features from the semantic decoder to focus on extracting details only within transition area. Specifically, the output feature from the last decoder block in the semantic decoder is used to generate spatial attention, since it is more related to semantics. Then, it goes through a max-pooling layer and average pooling layer along the channel axis, respectively. The pooled features are concatenated and go through a convolutional layer and a sigmoid layer to generate a spatial attention map. We use it to guide the matting decoder to attend to the transition area via an element-wise production operation and an element-wise sum operation. 

Given the predicted unified representation $U$ and guided matting result $M$ from the semantic decoder and matting decoder, we can derive the final alpha matte $\alpha$ as follows:
\begin{equation}
\alpha = (1-2\times \left | U -0.5 \right | )\times M + 2\times \left | U -0.5 \right | \times U.
\label{equa:hardattention}
\end{equation}
We adopt the commonly used alpha loss~\cite{xu2017deep} and Laplacian loss~\cite{hou2019context} on the predicted alpha matte $M$ and the final alpha matte $\alpha$ to supervise the matting decoder. Besides, we also use the composition loss~\cite{xu2017deep} on the final alpha matte $\alpha$.

\section{Experiment}

\subsection{Benchmark: Automatic Image Matting-500}

\begin{table}[b]
\footnotesize

\setlength{\tabcolsep}{2mm}{
\begin{tabular}{lcccccccc}

\hline
Dataset & Volume & Natural & SO & STM & NS & Class\\
\hline
 DAPM & 200 & $\surd$ & 200 & 0  & 0 & Portrait\\
 Comp-1k& 50 & $\times$&28 & 17 & 5 & Mixed\\
 HAtt& 50 & $\times$ & 30 & 11 & 9 & Mixed\\
 AM-2k & 200 & $\surd$ & 200 & 0  & 0 & Animal \\
 \hline
 AIM-500 & 500 & $\surd$ & 424 & 43  & 33 & Mixed\\
 \hline
\end{tabular}}
\caption{Comparison between AIM-500 with other matting test set.}
\label{tab:datasetcmoparison}
\end{table}

\begin{figure}[t]
    \includegraphics[width = 1\linewidth]{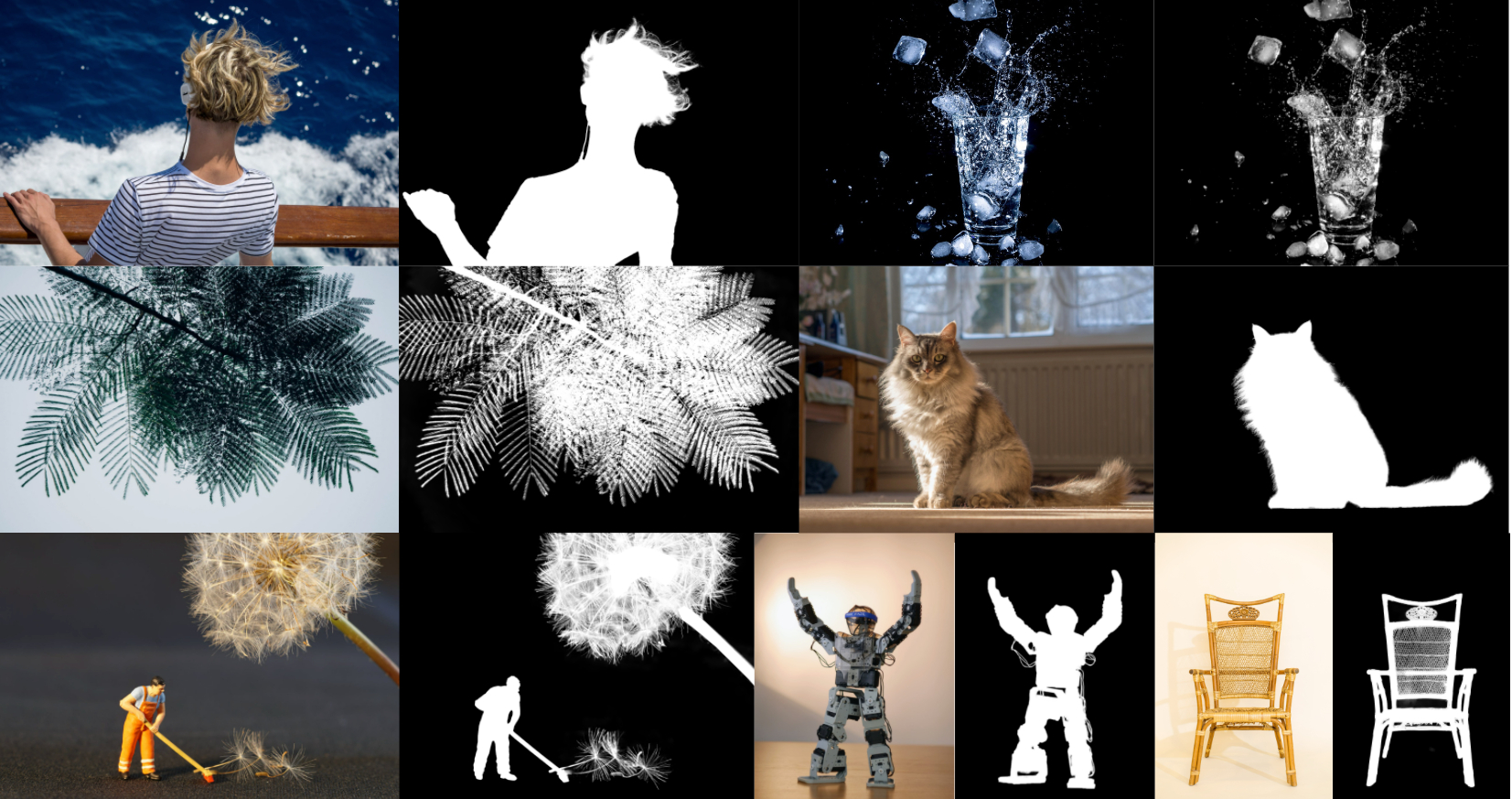}
    \caption{Some examples from our AIM-500 benchmark.}
    \label{fig:dataset}
\end{figure}

As discussed in Section~\ref{sec:rethinking}, previous work~\cite{zhang2019late,Qiao_2020_CVPR} evaluate their models either on synthetic test set such as Comp-1k~\cite{xu2017deep} or in-house test set of natural images limited to specific types, e.g., human and animal images~\cite{gfm,chen2018semantic}. In this paper, we establish the first natural image matting test set AIM-500, which contains 500 high-resolution real-world natural images from all three types and many categories. We collect the images from free-license websites and manually label the alpha mattes with professional software. The shorter side of each image is at least 1080 pixels. In Table~\ref{tab:datasetcmoparison}, we compare AIM-500 with other matting test sets including DAPM~\cite{shen2016deep}, Comp-1k~\cite{xu2017deep}, HAtt~\cite{Qiao_2020_CVPR}, and AM-2k~\cite{gfm} in terms of the volume, whether or not do they provide natural original images, the amount of three types images, and the object classes. We can see that AIM-500 is larger and more diverse than others, making it suitable for benchmarking AIM models. AIM-500 contains 100 portrait images, 200 animal images, 34 images with transparent objects, 75 plant images, 45 furniture images, 36 toy images, and 10 fruit images. We present some examples and their alpha mattes in Figure~\ref{fig:dataset}. Note that due to the privacy concern, all the portrait images have no identifiable information and are ready to release.

\begin{table*}[!t]
\resizebox{\linewidth}{!}{
\begin{tabular}{l|ccccc|c|cccc|cccccccc}

\hline
 & \multicolumn{5}{c}{Whole Image}  & \multicolumn{1}{|c}{Tran.}  & \multicolumn{4}{|c}{SAD-Type} & \multicolumn{8}{|c}{SAD-Category}\\
\hline
 & SAD  & MSE & MAD & Conn. & Grad. & SAD & SO & STM & NS & \multicolumn{1}{c|}{Avg.}  & Animal & Human & Transp. & Plant & Furni. & Toy & Fruit & Avg. \\
\hline
U2NET & 83.46 & 0.0348 & 0.0493 & 82.14 & 51.02 & \multicolumn{1}{c|}{43.37} & 69.69 & 120.59  & 211.98 & \multicolumn{1}{c|}{134.09}  & 67.67 & 89.50 & 210.34 & 75.72 & 87.20 & 54.64 & 52.24 & 91.04 \\
SHM & 170.44 & 0.0921 & 0.1012 & 170.67 & 115.29 & \multicolumn{1}{c|}{69.41} & 154.56 &204.67 &  329.9&\multicolumn{1}{c|}{229.71 }  & 174.65 & 141.49 &  333.24 & 157.24 &166.81  & 126.04 & 97.31 & 170.97 \\
LF & 191.74 & 0.0667 & 0.1130 & 181.26 & 63.51 & \multicolumn{1}{c|}{78.13} & 177.98 & 220.22 &331.34 &  \multicolumn{1}{c|}{243.18} & 167.90 & 131.96 & 276.13 &  228.94&  249.70& 224.50& 287.40  & 223.79 \\
HATT & 479.17 & 0.2700 & 0.2806 & 473.98 & 238.63 & \multicolumn{1}{c|}{114.23} &  509.75 & 338.11 &270.07 & \multicolumn{1}{c|}{372.64}  &  579.96&  484.85& 264.35 &433.96 &299.19 & 447.01 & 401.73 & 415.86\\

GFM & 52.66 & 0.0213 & 0.0313 & 52.69 & 46.11 & \multicolumn{1}{c|}{\underline{37.43}}  & 35.45 & 123.15 &\underline{181.90} & \multicolumn{1}{c|}{\underline{113.50}}  & 28.18 & 27.61 & \underline{190.50} & \underline{75.77} & 80.94 & \underline{51.42} & \underline{27.87} & \underline{68.90}\\
\hline
DIM$^*$ & \underline{49.27} & \textbf{0.0147} & \underline{0.0293} & \underline{47.10} & \textbf{29.30} &\multicolumn{1}{c|}{49.27} &  \textbf{19.51} & \underline{115.42} &345.33& \multicolumn{1}{c|}{160.09}  & \textbf{16.41} & \textbf{15.10} & 273.96 & 95.60 & \textbf{43.06} & \textbf{34.67} & \textbf{17.00} & 70.83 \\

\hline
Ours & \textbf{43.92}  & \underline{0.0161} & \textbf{0.0262} & \textbf{43.18} & \underline{33.05} & \multicolumn{1}{c|}{\textbf{30.74}}  & \underline{31.80} & \textbf{94.02} & \textbf{134.31} &\multicolumn{1}{c|}{\textbf{86.71}} & \underline{26.39} & \underline{24.68} & \textbf{148.68} & \textbf{54.03} & \underline{62.70} & 53.15 & 37.17 & \textbf{58.11}\\
\hline
\end{tabular}}
\caption{Quantitative results on AIM-500. DIM$^*$ denotes the DIM method using ground truth trimap as an extra input. Tran.: Transition Area, Transp.: Transparent, Furni.: Furniture.}
\label{tab:exp_results}
\end{table*}

\subsection{Implementation Details}

We trained our model and other representative matting models on the combination of matting datasets Comp-1k~\cite{xu2017deep}, HAtt~\cite{Qiao_2020_CVPR} and AM-2k~\cite{gfm}. To reduce the domain gap issue, we adopted the high-resolution background dataset BG-20k and the composition route RSSN proposed in \cite{gfm} to generate the training data. We composited each foreground image from Composition-1k and HAtt with five background images and each foreground image from AM-2k with two background images. The total number of training images is 8,710. To achieve better performance, we also adopted a type-wise data augmentation and transfer learning strategy during training.

\paragraph{Type-wise data augmentation.}After inspecting real-world natural images, we observed that NS images usually have a bokeh effect on the background. To simulate this effect, for all NS images, we added the blur effect on the background as RSSN to make the foreground prominent.

\paragraph{Transfer learning.}Although data augmentation could increase the number of training data, the number of original foreground images and their classes are indeed small. To mitigate the issue, we leveraged the salient object detection dataset DUTS~\cite{duts} for training since it contains more real-world images and classes. However, since images in DUTS have small resolutions (about $300\times400$) and little fine details, we only used it for pre-training and adopted a transfer learning strategy to finetune the pre-trained model further on the above synthetic matting dataset.

We trained our model on a single NVIDIA Tesla V100 GPU with batch size as 16, Adam as optimizer. When pre-training on DUTS, we resized all images to $320\times320$, set learning rate as $1\times10^{-4}$, trained for 100 epochs. During finetuning on the synthetic matting dataset, we randomly crop a patch with a size in $\{640\times640, 960\times960, 1280\times1280\}$ from each image and resized it to $320\times320$, set the learning rate as $1\times10^{-6}$, and trained for 50 epochs. It took about 1.5 days to train the model. We adopted the hybrid test strategy~\cite{gfm} with the scale factors 1/3 and 1/4, respectively.

\begin{figure*}[htb!]
    \includegraphics[width = 1\linewidth]{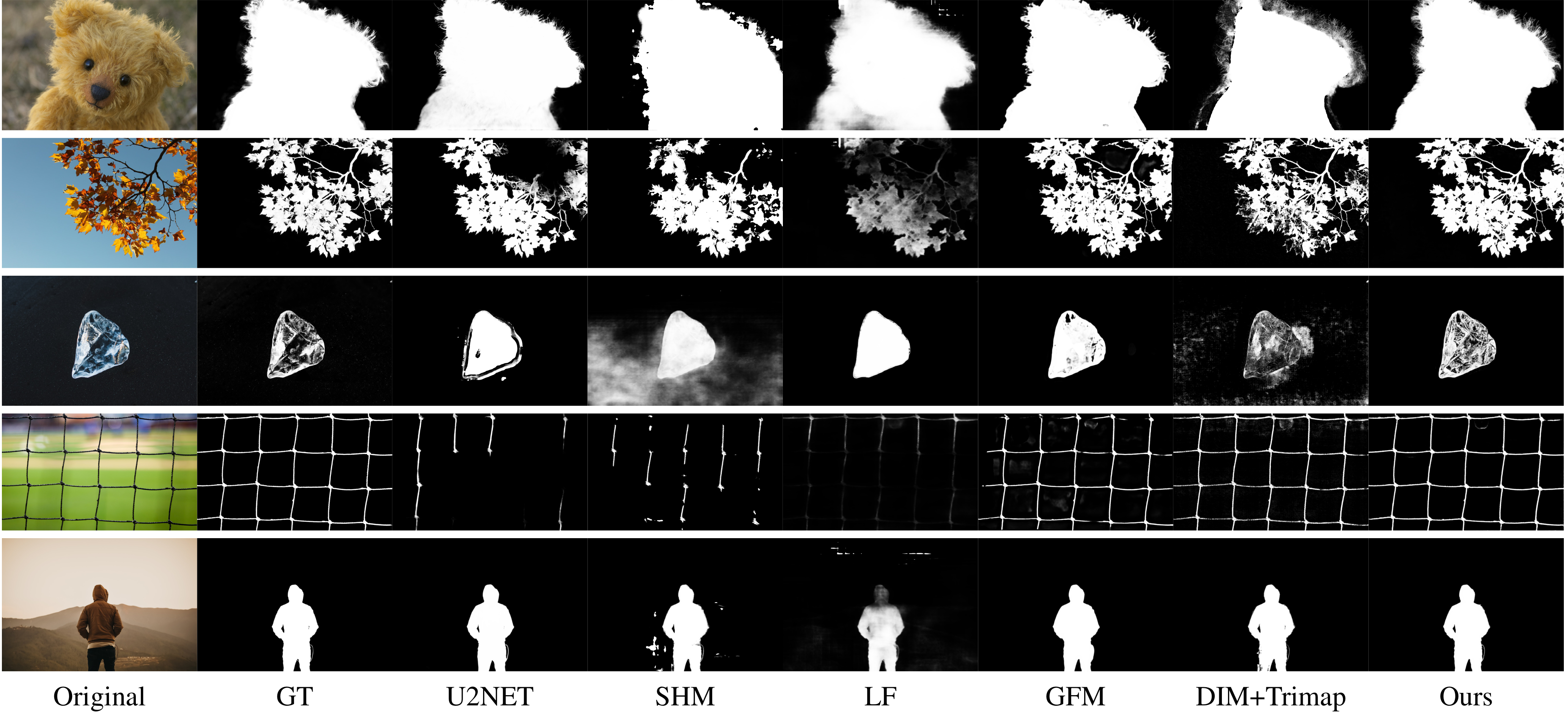}
    \caption{Some visual results of different methods on AIM-500. More results can be found in the supplementary material.}
    \label{fig:experiment_visual}
\end{figure*}

\subsection{Objective and Subjective Results}

We compare our model with several state-of-the-art matting methods including SHM~\cite{chen2018semantic}, LF~\cite{Qiao_2020_CVPR}, HAtt~\cite{Qiao_2020_CVPR}, GFM~\cite{gfm}, DIM~\cite{xu2017deep}, and an salient object detection method U2NET~\cite{u2net} on AIM-500. For LF, U2NET, and GFM, we used the code provided by the author. For SHM, HAtt, and DIM, we re-implemented the code since they are not available. Since DIM required a trimap for training, we used the ground truth trimap as the auxiliary input and denoted it as DIM$^*$. We adopted the same transfer learning strategy while training all the models. We chose the commonly used sum of absolute differences (SAD), mean squared error (MSE), Mean Absolute Difference (MAD), Connectivity(Conn.), Gradient(Grad.)~\cite{rhemann2009perceptually}, as the main metrics, and also calculated the SAD within transition areas, SAD by type and category for comprehensive evaluation. The results of objective metrics are summarized in Table~\ref{tab:exp_results}. Some visual results are presented in Figure~\ref{fig:experiment_visual}\footnote{More results can be found in the supplementary material.}. 

As shown in Table~\ref{tab:exp_results}, our model achieves the best in all metrics among all AIM methods and outperforms DIM$^*$ in most of the metrics. Although DIM$^*$ performs better in some SO category, e.g., animal, human, furniture, it indeed uses ground truth trimaps as auxiliary inputs while our model does not have such a requirement. Nevertheless, our model still outperforms DIM$^*$ in the transition areas as well as on the STM and NS types, implying that the semantic decoder predicts accurate semantic representation and the matting decoder has a better ability for extracting alpha details. Besides, DIM is very sensitive to the size of trimap and may produce a bad result when there is a large transition area, e.g. boundary of the bear in Figure~\ref{fig:experiment_visual}. U2NET can estimate the rough foregrounds for SO images, but it fails to handle STM and NS images, e.g. the crystal stone and the net, implying that a single decoder is overwhelmed to learn both global semantic features and local detail features to deal with all types of images since they have different characteristics. 

SHM uses a two-stage network to predict trimap and the final alpha, which may accumulate the semantic error and mislead the subsequent matting process. Consequently, it obtains large SAD errors in the whole image as well as in the transition area, i.e., 170.44 and 69.41. Some failure results can be found in the bear and crystal stone in Figure~\ref{fig:experiment_visual}. LF adopts a classification network to distinguish foreground and background, but it is difficult to adapt to STM and NS images as they do not have explicit foreground and background, thereby resulting in large average SAD errors. HATT tries to learn foreground profile to support boundary detail matting. However, there are no explicit supervisory signal for the semantic branch, which makes it difficult to learn explicit semantic representations. As a strong baseline model, GFM outperforms other AIM methods, but it is still worse than us, i.e., 52.66 to 43.92, especially for STM and NS images, as seen from the crystal stone and the net. The results demonstrate that the customized designs in the network matter a lot for dealing different types of images. Generally, our method achieves the best performance both objectively and subjectively.

\subsection{Ablation Study}
\label{sec:ablation}

\begin{table}[t]
\begin{center}
\setlength{\tabcolsep}{2.0mm}{
\begin{tabular}{ccccc|ccc}
\hline
UNI & TL & MP & SE & SA & SAD & MSE & MAD \\
\hline
&& & &  & 81.08 & 0.0363& 0.0480\\
$\checkmark$ && & &  & 76.90 & 0.0296 & 0.0456 \\
$\checkmark$ &$\checkmark$ & & & & 52.66 & 0.0213 & 0.0313\\
$\checkmark$ &$\checkmark$ &$\checkmark$ & & & 51.23 &0.0205 &0.0307 \\
$\checkmark$ &$\checkmark$&  & $\checkmark$ & & 48.52 & 0.0195& 0.0287\\
$\checkmark$ &$\checkmark$&  &   & $\checkmark$ &48.95&0.0207&0.0293\\
$\checkmark$ &$\checkmark$ & & $\checkmark$ & $\checkmark$& 44.50 &\textbf{0.0158} & 0.0262\\
$\checkmark$ &$\checkmark$ &$\checkmark$ & $\checkmark$ & $\checkmark$  & \textbf{43.92}  & 0.0161 &\textbf{ 0.0262} \\
 \hline
\end{tabular}}
\end{center}
\caption{Ablation study results. UNI: unified semantic representations; TL: transfer learning; MP: backbone with max pooling; SE: SE attention; SA: spatial attention.}
\label{tab:ablation}
\end{table}

We present the ablation study results in Table~\ref{tab:ablation}. There are several findings. Firstly, the proposed unified semantic representation is more effective than the traditional trimap, which  helps to reduce the SAD from 81.08 to 76.90. Secondly, the transfer learning strategy is very effective by leveraging the large scale DUTS dataset for pre-training, reducing the SAD from 75.90 to 52.66. Thirdly, all the customized designs including max-pooling in the backbone, SE attention module in the semantic decoder, and spatial attention based on semantic features to guide the matting decoder, are useful and complementary to each other. For example, our method that uses all these designs reduces SAD from 52.66 to 43.92. It provides explicit and non-trivial supervisory signals for the semantic decoder, facilitating it to learn effective semantic features. It is noteworthy that although our model without MP has lower MSE, however, the visual results of the alpha matte obtained by our model with MP is better, e.g., with clearer details. Thereby, we choose it as our final model.

\section{Conclusion}

In this paper, we investigate the difficulties in automatic image matting, including matting without auxiliary inputs, matting on natural images, and matting on all types of images. We propose a unified semantic representation for all types by introducing the new concepts of duomap and unimap, proved to be useful. We devise an automatic matting network with several customized new designs to improve its capability for AIM. Moreover, we establish the first natural image test set AIM-500 to benchmark AIM models, which can serve as a test bed to facilitate future research. 

\bibliographystyle{named}
\bibliography{matting}

\end{document}


\maketitle

In this supplementary material, we present the network structure detail of our model, more details about the proposed natural image matting test set AIM-500, more visual results of our model and state-of-the-art methods on AIM-500, and more analysis of the ablation study.

\section{More details about our model}

\subsection{Network structure}

Table~\ref{Tab:network} shows the detailed structure of some essential parts in our network including the shared encoder, the semantic decoder based on the unified semantic representation, and the guided matting decoder. Our network has 55.3M parameters and 191.52GMac computational complexity. It takes about 0.1633s for the network to process a 800$\times$800 image.

\section{More details about AIM-500}

\subsection{Compare with synthetic validation set}
As we mentioned in the paper, due to the unavailability of natural test with labeled alpha mattes, previous AIM methods evaluate their results on the synthetic test set, which is generated by pasting high-resolution foregrounds from matting dataset~\cite{xu2017deep,Qiao_2020_CVPR} on backgrounds from low-resolution COCO dataset~\cite{lin2014microsoft} or PASCAL VOC~\cite{everingham2010pascal}. 

\begin{figure*}[htb!]
    \includegraphics[width = 1\linewidth]{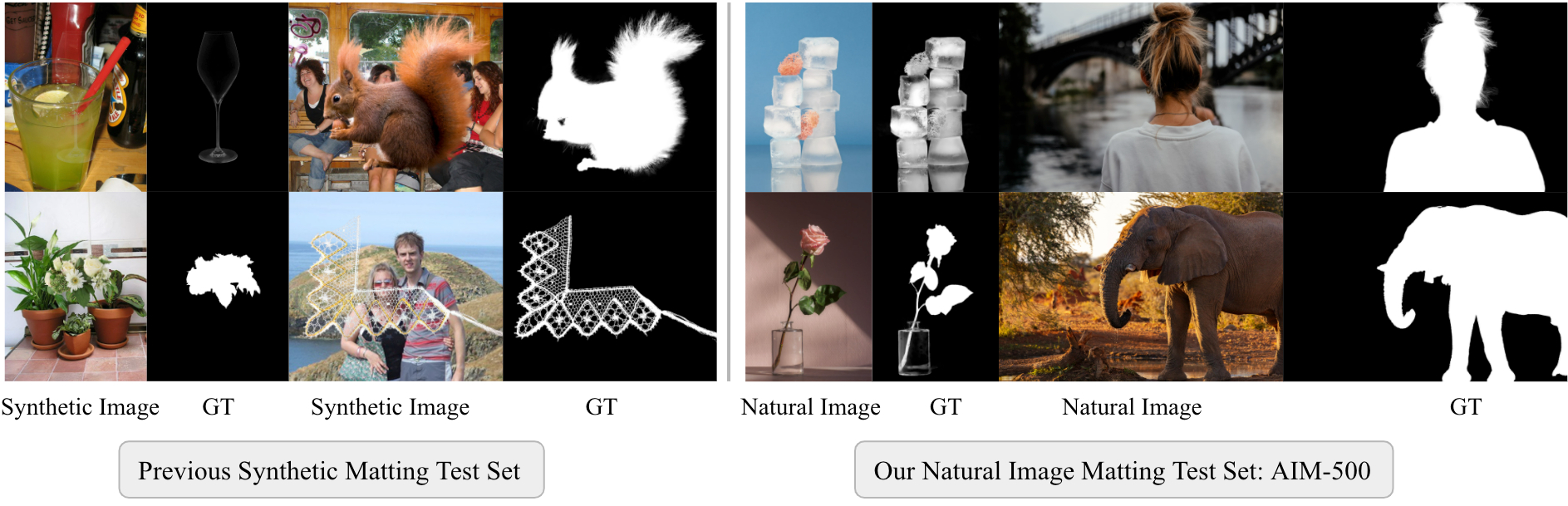}
    \caption{Comparison between our proposed natural images matting test set AIM-500 and previous synthetic matting test set.}
    \label{fig:compare}
\end{figure*}

However, as can be seen from Figure~\ref{fig:compare}, such procedure results in large bias with real-world natural images and wrong label because of the two reasons, 1) Pasting other salient objects on backgrounds will make the original ground truth unreasonable, for examples, pasting a transparent cup in front of another cup or pasting a squirrel in front of a lot of people. Such cases appear a lot in the synthetic matting test set and result in the biased evaluation of AIM models; and 2) Synthetic images have non-negligible composite artifacts, performing well on such a test set does not necessarily imply a good generalization ability on real-world images.

In contrast to them, the proposed natural image matting test set named AIM-500 contains natural images rather than composite images with manually labeled alpha mattes as shown in Figure~\ref{fig:compare}. We believe AIM-500 can serve as a good test bed to benchmark AIM models.

\subsection{More examples of AIM-500}
\begin{figure}[htb]
    \includegraphics[width = 1\linewidth]{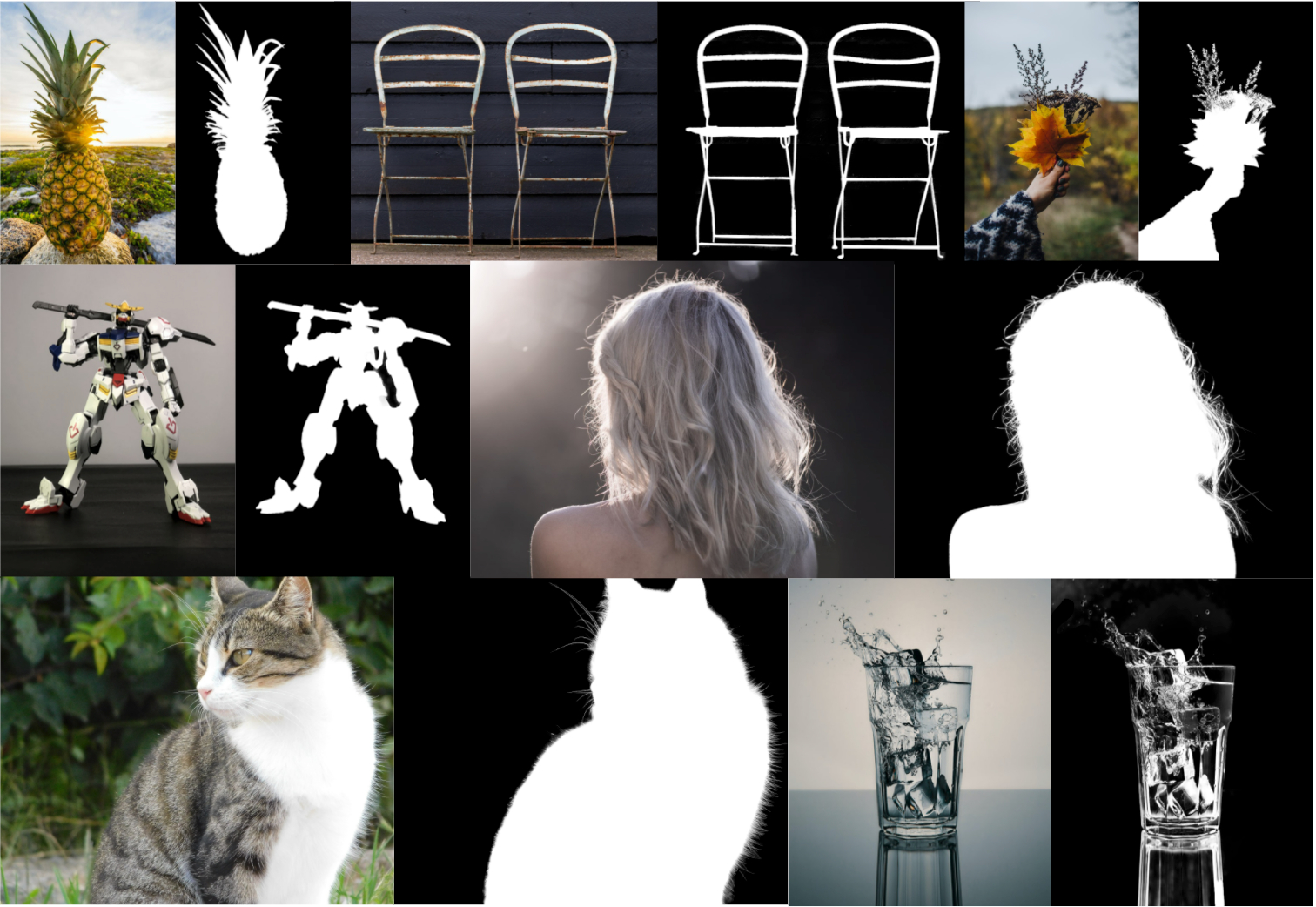}
    \caption{More examples from the proposed AIM-500.}
    \label{fig:example}
\end{figure}

AIM-500 contains 7 categories including portrait, animal, fruit, furniture, toy, plant, and transparent objects. Here we show more examples of each category in Figure~\ref{fig:example}. As can be seen, images in AIM-500 have diverse foregrounds and backgrounds. 

\section{More results of Experiment}

\subsection{More visual results on AIM-500}

\begin{figure*}[htb!]
    \includegraphics[width = 1\linewidth]{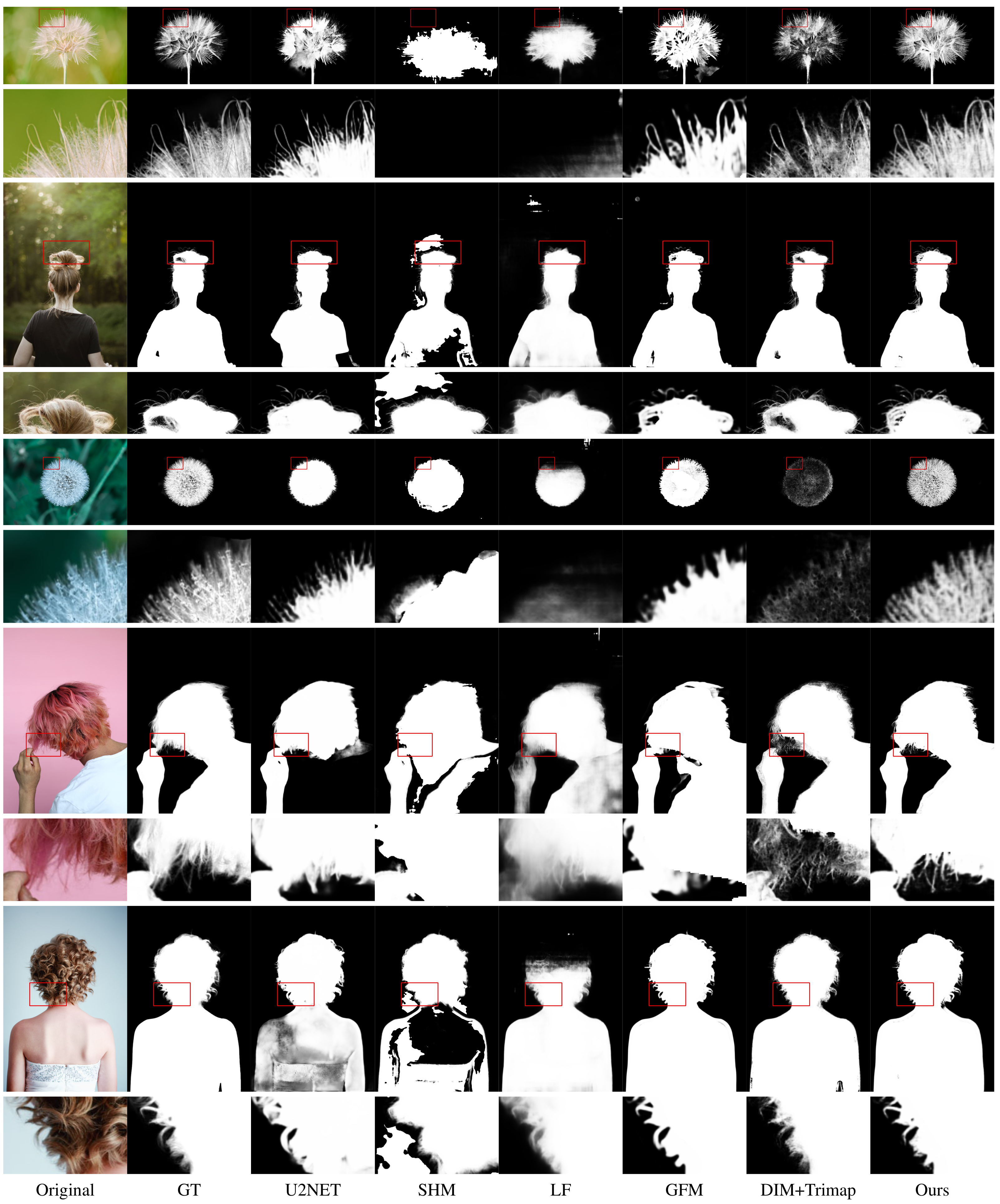}
    \caption{Additional visual results on AIM-500.}
    \label{fig:experiment_visual1}
\end{figure*}

\begin{figure*}[htb!]
    \includegraphics[width = 1\linewidth]{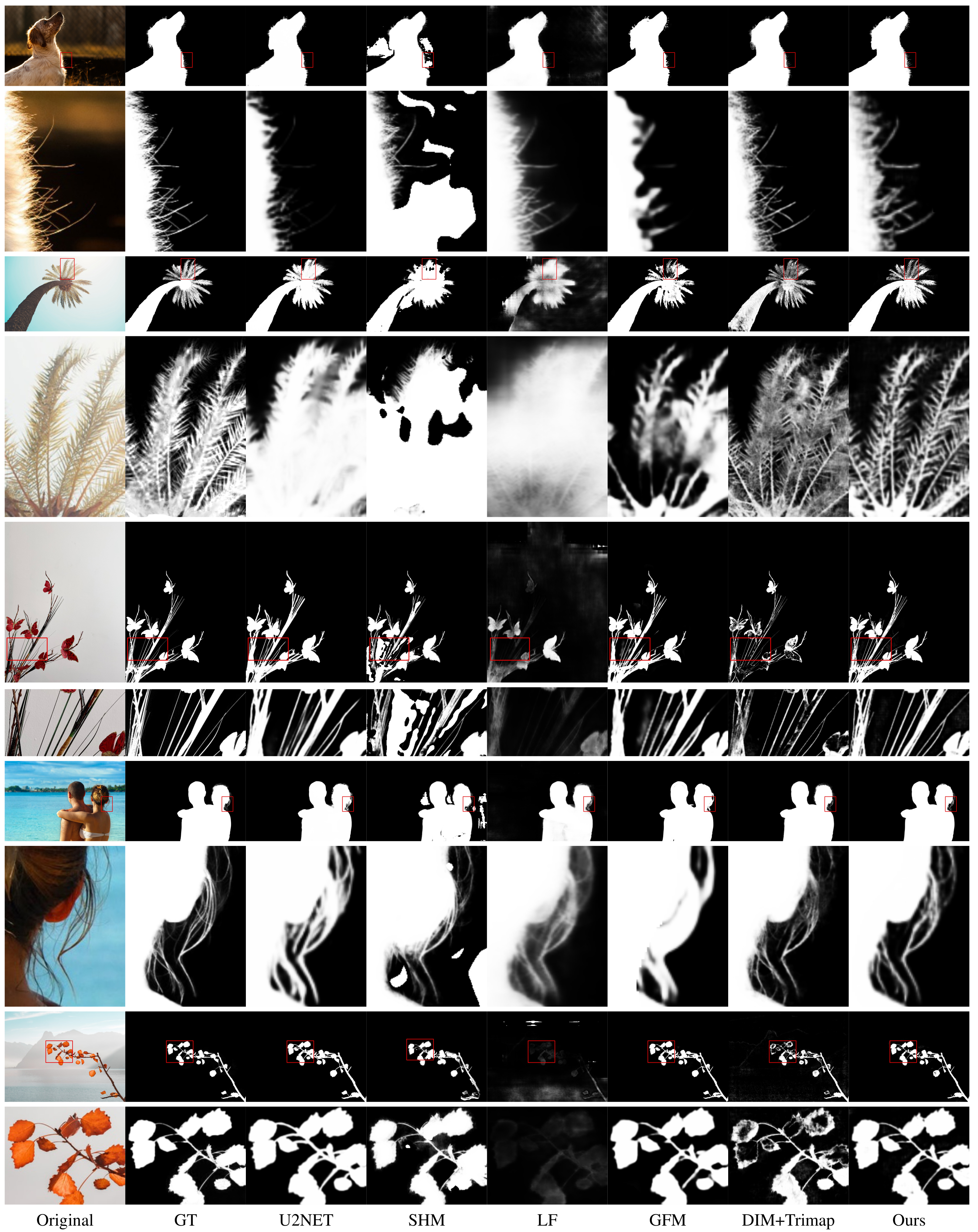}
    \caption{Additional visual results on AIM-500.}
    \label{fig:experiment_visual2}
\end{figure*}

In Figure~\ref{fig:experiment_visual1} and Figure~\ref{fig:experiment_visual2}, we compare our model with several state-of-the-art methods U2NET~\cite{u2net}, SHM~\cite{chen2018semantic}, LF~\cite{zhang2019late}, GFM~\cite{gfm}, and DIM+Trimap~\cite{xu2017deep} on AIM-500. It can be seen that our method achieves the best results on all three types of images. U2Net and GFM fail to extract the fine detail. SHM and LF have many errors on semantic background or foreground. Even with extra ground truth trimap as input, DIM performs worse than us especially when the trimap region is large, e.g., SO images. 

\subsection{More of ablation study}

\begin{figure}[htb!]
    \includegraphics[width = 1\linewidth]{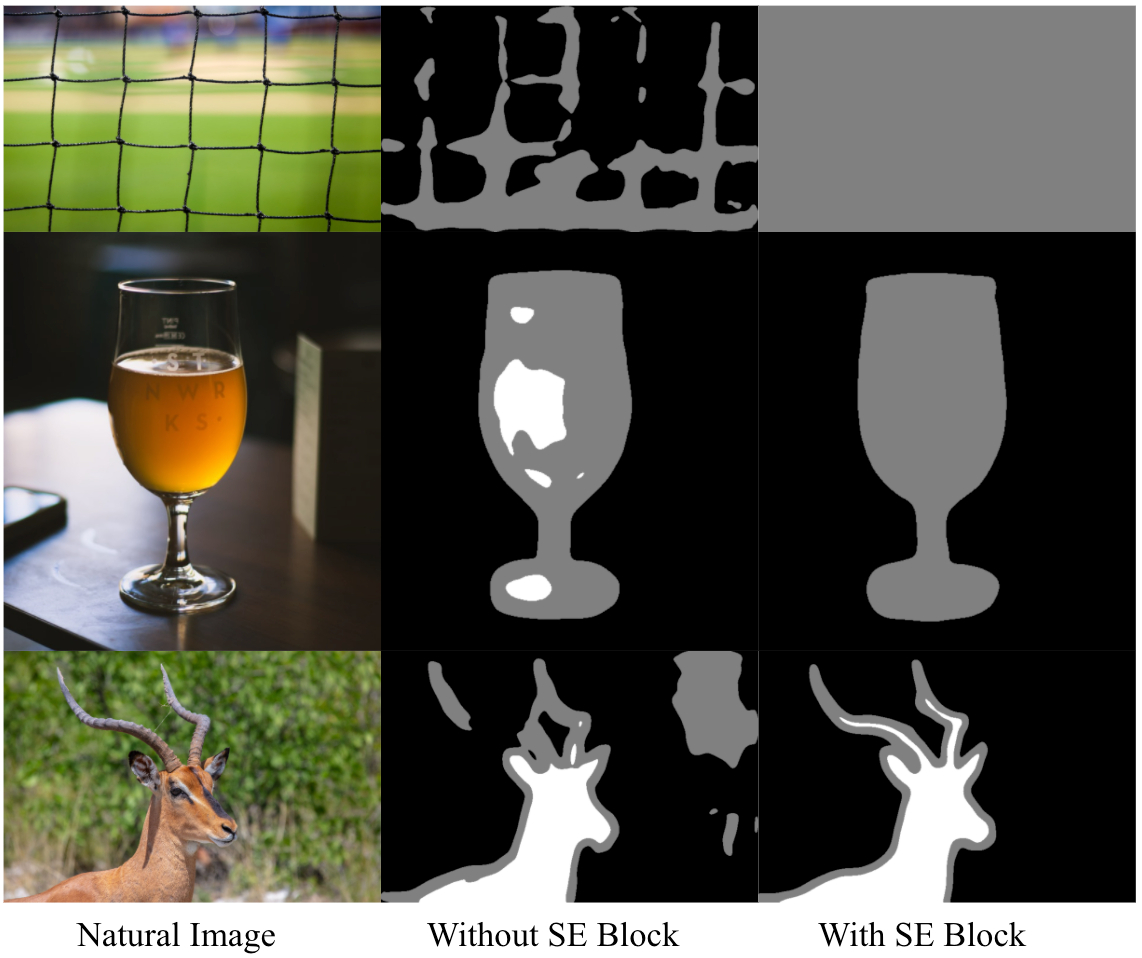}
    \caption{Visual results of the predicted unified semantic representations with or without SE blocks in the semantic decoder.}
    \label{fig:seblock}
\end{figure}

\textbf{SE attention block} We adopted five Squeeze-and-Excitation attention modules~\cite{hu2018squeeze} in the semantic decoder to re-calibrate the features and select the most informative features for predicting unified representations. Here we provide objective and subjective results of it. In Table~\ref{tab:ablation}, we show the results of IOU (Intersection over Union) and the accuracy of the predicted unified semantic representations by the semantic decoder with or without SE blocks. We also show some visual results in Figure~\ref{fig:seblock}. As can be seen, for all three types of images, SE blocks benefit the semantic decoder to predict better unified semantic representations.

\begin{table}[htb]
\begin{center}

\begin{tabular}{c|ccc}
\hline
SE Blocks & IOU & Accuracy \\
\hline
$\times$  & 0.7002 & 0.8756 \\
$\checkmark$ & 0.7393 & 0.8900 \\

 \hline
\end{tabular}
\end{center}
\caption{IOU and accuracy of the predicted unified semantic representations with or without SE blocks in the semantic decoder.}
\label{tab:ablation}
\end{table}

\clearpage

\bibliographystyle{named}
\bibliography{supp}

\newpage

\begin{table*}
\centering

\footnotesize

\begin{tabular}{|c|c|c|}
\hline
Block & Output size & Detail  \\
\hline

\hline
\multicolumn{3}{|c|}{\textbf{Improved Backbone}}  \\
\hline
$E_0$ & $N\times64\times320\times320$ & conv ($7\times7$, 64, stride 1, padding 3) + BN + ReLU\\
\hline
$M_0$ & $N\times64\times160\times160$ & maxpool ($3\times3$, stride 2, padding 1, return indices)\\
\hline
$M_1$ & $N\times64\times80\times80$ & maxpool ($3\times3$, stride 2, padding 1, return indices)\\
\hline
$E_1$ & $N\times64\times80\times80$ & ResNet-34~\cite{he2016deep} layer 1 (stride 1)\\
\hline
$M_2$ & $N\times64\times40\times40$ & maxpool ($3\times3$, stride 2, padding 1, return indices)\\
\hline
$E_2$ & $N\times128\times40\times40$ & ResNet-34~\cite{he2016deep} layer 2 (stride 1)\\
\hline
$M_3$ & $N\times128\times20\times20$ & maxpool ($3\times3$, stride 2, padding 1, return indices)\\
\hline
$E_3$ & $N\times256\times20\times20$ & ResNet-34~\cite{he2016deep} layer 3 (stride 1)\\
\hline
$M_4$ & $N\times256\times10\times10$ & maxpool ($3\times3$, stride 2, padding 1, return indices)\\
\hline
$E_4$ & $N\times512\times10\times10$ & ResNet-34~\cite{he2016deep} layer 4 (stride 1)\\
\hline

\hline
\multicolumn{3}{|c|}{\textbf{Unified Semantic Decoder }}  \\
\hline
$PPM$ & $N\times512\times10\times10$ & PSPModule~\cite{zhao2017pyramid} (512, multiscale=1,3,5)\\
\hline
$SD_4$ & $N\times256\times20\times20$ & \tabincell{c}{conv ($3\times3$, 512, stride 1) + BN + ReLU\\ conv ($3\times3$, 512, stride 1) + BN + ReLU\\conv ($3\times3$, 256, stride 1) + BN + ReLU\\ upsample(2)}\\
\hline
$SE_4$  & $N\times256\times20\times20$ & avgpool + linear(256, 16) + ReLU + linear(16, 256) + Sigmoid\\
\hline
$SD_3$ & $N\times128\times40\times40$ & \tabincell{c}{conv ($3\times3$, 256, stride 1) + BN + ReLU\\ conv ($3\times3$, 256, stride 1) + BN + ReLU\\conv ($3\times3$, 128, stride 1) + BN + ReLU\\ upsample(2)}\\
\hline
$SE_3$  & $N\times128\times40\times40$ & avgpool + linear(128, 8) + ReLU + linear(8, 128) + Sigmoid\\
\hline
$SD_2$ & $N\times64\times80\times80$ & \tabincell{c}{conv ($3\times3$, 128, stride 1) + BN + ReLU\\ conv ($3\times3$, 128, stride 1) + BN + ReLU\\conv ($3\times3$, 64, stride 1) + BN + ReLU\\ upsample(2)}\\
\hline
$SE_2$  & $N\times64\times80\times80$ & avgpool + linear(64, 4) + ReLU + linear(4, 64) + Sigmoid\\
\hline
$SD_1$ & $N\times64\times160\times160$ & \tabincell{c}{conv ($3\times3$, 64, stride 1) + BN + ReLU\\ conv ($3\times3$, 64, stride 1) + BN + ReLU\\conv ($3\times3$, 64, stride 1) + BN + ReLU\\ upsample(2)}\\
\hline
$SE_1$  & $N\times64\times160\times160$ & avgpool + linear(64, 4) + ReLU + linear(4, 64) + Sigmoid\\
\hline
$SD_0$ & $N\times64\times320\times320$ & \tabincell{c}{conv ($3\times3$, 64, stride 1) + BN + ReLU\\ conv ($3\times3$, 64, stride 1) + BN + ReLU\\ upsample(2)}\\
\hline
$SPA$  & $N\times1\times320\times320$ & maxpool + avgpool + conv ($7\times7$, 1, padding 3)\\
\hline
$SE_0$  & $N\times64\times320\times320$ & avgpool + linear(64, 4) + ReLU + linear(4, 64) + Sigmoid\\
\hline
$SD-final$ & $N\times3\times320\times320$ & conv ($3\times3$, 3, stride 1, padding 1)  \\
\hline

\hline
\multicolumn{3}{|c|}{\textbf{Semantic Guided Matting Decoder}}  \\
\hline
$MD_5$ & $N\times512\times10\times10$ & \tabincell{c}{conv ($3\times3$, 512, dilation 2, padding 2) + BN + ReLU\\ conv ($3\times3$, 512, dilation 2, padding 2) + BN + ReLU\\conv ($3\times3$, 512, dilation 2, padding 2) + BN + ReLU}\\
\hline
$MD_4$ & $N\times256\times10\times10$ & \tabincell{c}{conv ($3\times3$, 512, stride 1) + BN + ReLU\\ conv ($3\times3$, 512, stride 1) + BN + ReLU\\conv ($3\times3$, 256, stride 1) + BN + ReLU}\\
\hline
$MU_4$ & $N\times256\times20\times20$ & max-unpool ($2\times2$, stride 2)\\
\hline
$MD_3$ & $N\times128\times20\times20$ & \tabincell{c}{conv ($3\times3$, 256, stride 1) + BN + ReLU\\ conv ($3\times3$, 256, stride 1) + BN + ReLU\\conv ($3\times3$, 128, stride 1) + BN + ReLU}\\
\hline
$MU_3$ & $N\times128\times40\times40$ & max-unpool ($2\times2$, stride 2)\\
\hline
$MD_2$ & $N\times64\times40\times40$ & \tabincell{c}{conv ($3\times3$, 128, stride 1) + BN + ReLU\\ conv ($3\times3$, 128, stride 1) + BN + ReLU\\conv ($3\times3$, 64, stride 1) + BN + ReLU}\\
\hline
$MU_2$ & $N\times64\times80\times80$ & max-unpool ($2\times2$, stride 2)\\
\hline
$MD_1$ & $N\times64\times80\times80$ & \tabincell{c}{conv ($3\times3$, 64, stride 1) + BN + ReLU\\ conv ($3\times3$, 64, stride 1) + BN + ReLU\\conv ($3\times3$, 64, stride 1) + BN + ReLU}\\
\hline
$MU_1$ & $N\times64\times160\times160$ & max-unpool ($2\times2$, stride 2)\\
\hline
$MU_0$ & $N\times64\times320\times320$ & max-unpool ($2\times2$, stride 2)\\
\hline
$MD_0$ & $N\times64\times320\times320$ & \tabincell{c}{conv ($3\times3$, 64, stride 1) + BN + ReLU\\ conv ($3\times3$, 64, stride 1) + BN + ReLU}\\
\hline
$SPAR$ & $N\times64\times320\times320$ & SPA + SPA $\times MD_0$ \\
\hline
$MD-final$ & $N\times1\times320\times320$ & conv ($3\times3$, 3, stride 1, padding 1)  \\
\hline

\hline
\multicolumn{3}{|c|}{\textbf{Matting Fusion}}  \\
\hline
$MF$ & $N\times1\times320\times320$ & pixel-wise sum and multiplication.\\
\hline

\end{tabular}
\caption{The network structure of our matting model. The input size is $N\times3\times320\times320$, where N is the batch size. }

\label{Tab:network}
\end{table*}

